\documentclass[final]{style/cvpr}
\usepackage{subcaption} 
\usepackage{times}
\usepackage{epsfig}
\usepackage{graphicx}
\usepackage{amsmath}
\usepackage{array}
\usepackage{graphicx}
\usepackage{amssymb}
\usepackage{multirow}
\usepackage{adjustbox}
\usepackage{tabularx}
\usepackage{xcolor,colortbl}
\usepackage{booktabs}
\usepackage{xspace}
\usepackage{url}

\usepackage{algorithmic}
\usepackage[ruled, linesnumbered, noend]{algorithm2e}
\usepackage{xcolor}

\SetCommentSty{mycommfont}

\usepackage{amssymb}
\usepackage{pifont}
\usepackage{soul}

\newcommand*\BitOr{\mathbin{|}}

\makeatletter
\DeclareRobustCommand\onedot{\futurelet\@let@token\@onedot}
\def\@onedot{\ifx\@let@token.\else.\null\fi\xspace}

\def\ie{\emph{i.e}\onedot}

\def\etal{\emph{et al}\onedot}

\DeclareMathOperator*{\argmax}{argmax}

\newcommand*{\rom}[1]{\expandafter\@slowromancap\romannumeral #1@}
\definecolor{Gray}{gray}{0.9}
\definecolor{GrayLine}{gray}{0.7}
\makeatother

\usepackage[pagebackref=true,breaklinks=true,colorlinks,bookmarks=false]{hyperref}

\def\cvprPaperID{10} 
\def\confYear{CVPR 2021}

\begin{document}

\title{$\text{E}^2$VTS: Energy-Efficient Video Text Spotting from Unmanned Aerial Vehicles}
\author{Zhenyu~Hu$^{1}$\footnotemark[1], Zhenyu~Wu$^{1}$\thanks{The first two authors contribute equally and are listed alphabetically.}, Pengcheng~Pi$^{1}$, Yunhe~Xue$^{1}$, Jiayi~Shen$^{1}$, Jianchao~Tan$^{2}$, Xiangru~Lian$^{2}$, \\ Zhangyang~Wang$^{3}$, and Ji~Liu$^{2}$ \thanks{Correspondence to: Ji Liu with the AI Platform, Ytech Seattle AI Lab, FeDA Lab, Kwai Inc., Seattle WA 98004, USA, e-mail: $\langle \text{ji.liu.uwisc@gmail.com} \rangle$ }\\
$^1$Texas A\&M University\hspace{1em} $^2$Kwai Inc.\hspace{1em} $^3$The University of Texas at Austin

}
\maketitle

\maketitle

\begin{abstract}
Unmanned Aerial Vehicles (UAVs) based video text spotting has been extensively used in civil and military domains. 
UAV's limited battery capacity motivates us to develop an energy-efficient video text spotting solution.  
In this paper, we first revisit RCNN's crop \& resize training strategy and empirically find that it outperforms aligned RoI sampling on a real-world video text dataset captured by UAV. To reduce energy consumption, we further propose a multi-stage image processor that takes videos' redundancy, continuity, and mixed degradation into account.
Lastly, the model is pruned and quantized before deployed on Raspberry Pi.
Our proposed energy-efficient video text spotting solution, dubbed as $\text{E}^2VTS$, outperforms all previous methods by achieving a competitive tradeoff between energy efficiency and performance. 
All our codes and pre-trained models are available at {\emph{\url{https://github.com/wuzhenyusjtu/LPCVC20-VideoTextSpotting}}}.
\end{abstract}


\section{Introduction} 
UAV-based video text spotting is broadly applied in assistive navigation, automatic translation, road sign recognition, industrial monitoring, and disaster response, etc. A standard video text spotting model has four components: text detector, text recognizer, text tracker, and post-processing.

Existing video text spotting solutions~\cite{wang2017end,cheng2020free} are purely performance-driven and fail to take energy consumption into account. Multi-frame related features are first obtained in frame-wise detection or tracking. Then, they are aggregated for enhancement in a cross-frame and multi-scale way for text recognition. Therefore, existing performance-driven solutions are high in energy consumption and unsuitable for resource-constrained UAV platforms.


In this paper, we propose an Energy-Efficient Video Text Spotting solution, dubbed as $\text{E}^2$VTS. Our contribution can be summarized as follows:

\begin{itemize}
    \item \textbf{Novel Training \& Inference Strategies:} To obtain better text spotting performance, we revisit RCNN and empirically find that crop and resize outperforms aligned RoI Pooling when connecting the text recognizer with the text detector. To further save energy consumption, we propose a multi-stage image processor to select the highest-quality frame in a sliding window, reject text-free frames as well as crop non-text regions, and reject out-of-distribution frames.
    
    \item \textbf{Experiments:} On a real-world UAV-captured text video dataset deployed on Raspberry Pi, we conducted thorough ablation studies on the proposed training and inference strategies. The evaluation metric takes both energy consumption and text spotting performance into consideration. Models are pruned and quantized before deployment. 
\end{itemize}
 
\section{Related Work}
\subsection{Text Reading in Images}
\noindent\textbf{Text Detection:}
Object detection-based~\cite{liao2018textboxes++,zhou2017east,ma2018arbitrary,zhang2019look,wang2019arbitrary} and sub-text components-based~\cite{deng2018pixellink,wu2017self,tian2019learning,wang2019shape,baek2019character} are two streams of solutions to text detection. Based on the observation that any part of a text instance is still text, sub-text components-based methods incorporate the inductive biases of the homogeneity and locality of text instances into model design.

\noindent\textbf{Text Recognition:}
There are two major strategies in decoding text content from image encoded features from CNN, Connectionist Temporal Classiﬁcation (CTC)~\cite{graves2006connectionist} and the encoder-decoder framework~\cite{sutskever2014sequence}. CTC-based methods includes CRNN~\cite{shi2016end} and Sliding Convolutional Character Models~\cite{yin2017scene}. Encoder-Decoder Methods include $\text{R}^2$AM~\cite{lee2016recursive} and Edit Probability (EP)~\cite{bai2018edit}.

\noindent\textbf{End-to-end Text Spotting:}
The representative end-to-end text spotting are summarized into two categories, regular-shaped and arbitrary-shaped.

\textit{Regular-shaped Text:}
Li~\etal~\cite{li2017towards, wang2019towards} proposed the first deep-learning based end-to-end trainable scene text spotting method for horizontal text by incorporating RoI Pooling~\cite{ren2016faster} to join the detection and recognition stage. 
Deep TextSpotter~\cite{busta2017deep} handled multi-orientation text instances without feature sharing between the detection and recognition stages.
%
End-to-End TextSpotter~\cite{he2018end} and FOTS~\cite{liu2018fots} adopted an anchor-free mechanism to improve both the training and inference speed. They use two similar sampling strategies, \ie, Text-Alignment and RoIRotate, to extract feature from arbitrary-oriented quadrilateral detection results.

\textit{Arbitrary-shaped Text:}
Mask TextSpotter~\cite{liao2019mask, liao2020mask} used character-level supervision to simultaneously detect and recognize characters and instance masks. Nonetheless, the character-level ground truths are expensive, thus mostly unavailable for real data. RoI Masking~\cite{qin2019towards} cropped out the features from the predicted axis-aligned rectangular bounding boxes and multiplied the features with the corresponding instance segmentation mask. 
TextDragon~\cite{feng2019textdragon} proposed RoISlide to transform the whole text features into axis-aligned features indirectly by transforming each local quadrangle sequentially. 
As the first one-stage text spotting method, CharNet~\cite{xing2019convolutional} directly outputted bounding boxes of words, with corresponding character labels. 
%
ABCNet~\cite{Liu_2020_CVPR} adaptively fitted arbitrarily-shaped text by a parameterized Bezier curve and used BezierAlign layer to extract accurate convolution features. 
CRAFTS~\cite{baek2020character} used the character region feature from the detector as input character attention to the recognizer.

\subsection{Text Reading in Videos} 

\noindent\textbf{Text Detection \& Tracking:} Wang~\etal~\cite{wang2019video} proposed a multi-scale feature sampling and warping network on adjacent frames, and an attention-based multi-frame feature aggregation mechanism to fuse the complementary text features from related frames.
Wu~\etal~\cite{wu2015new} explored Delaunay triangulation to detect and track texts. The triangular mesh pattern reflects text properties, such as regular spacing between characters and constant stroke width, thus distinguishable from non-text.
Yang~\etal~\cite{yang2017unified} combined single-frame detection with cross-frame motion-based tracking. The text association was formulated into a cost-flow network.
Tian~\etal~\cite{tian2016scene} located character candidates
locally and searched text regions globally. Specifically, a multi-strategy tracking based text detection approach~\cite{zuo2015multi} was used to globally search and select the best text region with dynamic programming.
Wang~\etal~\cite{wang2019ICME} proposed a fully convolutional model based on a novel refine block structure, which refines the low-resolution semantic features with the high-resolution low-level features.



\noindent\textbf{End-to-End Text Spotting}
Wang~\etal~\cite{wang2017end} proposed a multi-frame tracking based method, where text detection and recognition are done on each frame before recognized texts are tracked over the video sequence.
FREE~\cite{cheng2020free,cheng2019you} proposed a text recommender to select the highest-quality text from text streams for recognizing and released a large scale video text spotting dataset.




\newcommand{\norm}[1]{\lvert #1 \rvert}
\section{$\text{E}^2$VTS: An Energy-Efficient Video Text Spotting Solution}

\paragraph{Overview:} 

\begin{figure}[h!]
  \centering
  \includegraphics[width=0.475\textwidth]{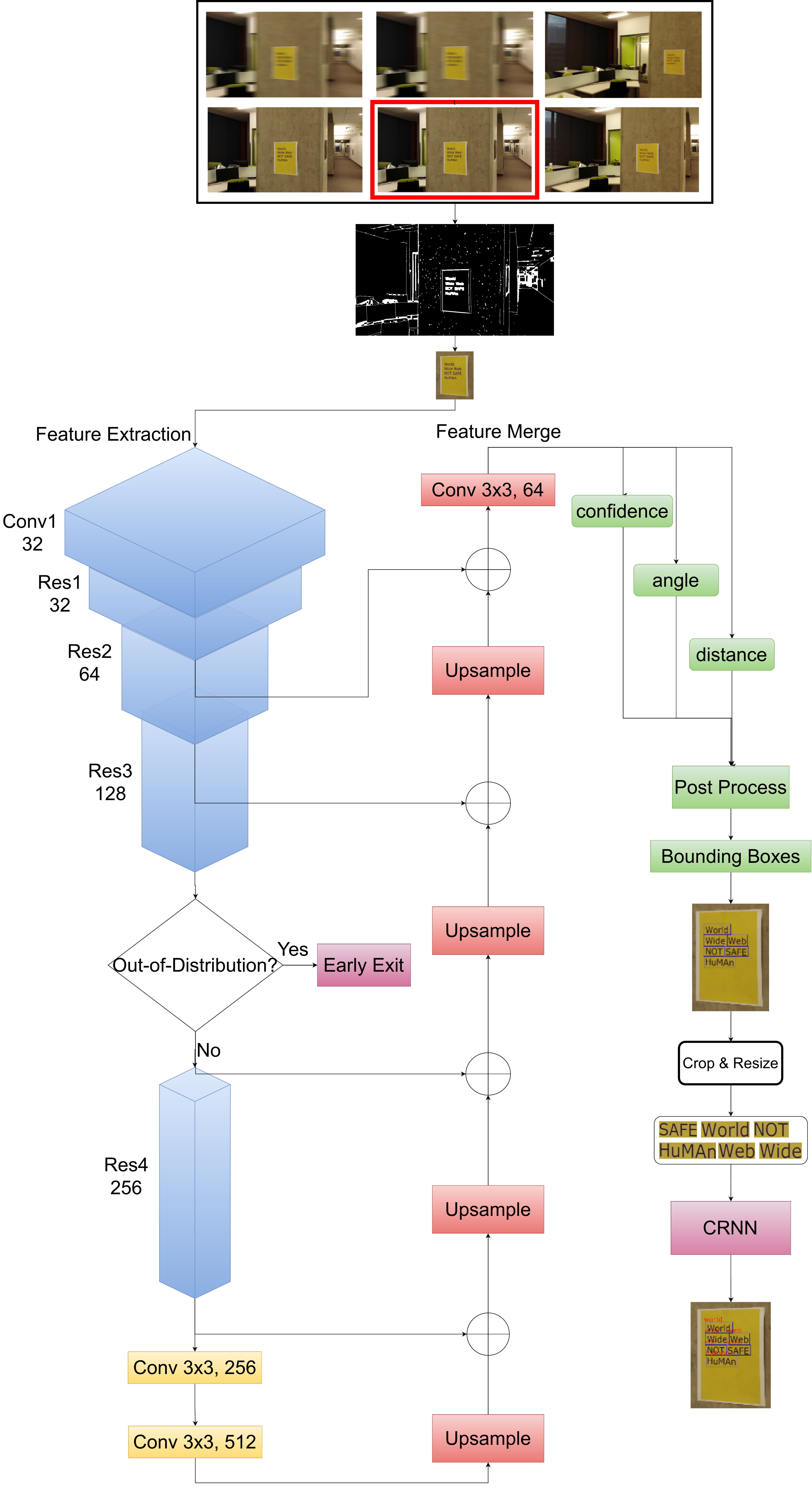}
  \caption{ Overview: $\text{E}^2$VTS consists of two components. Component one is a multi-stage image processor which selects the best frame within a window size and crops out the background. Component two is a two-step crop \& resize text spotting system including an EAST detector and a CRNN recognizer. The EAST detector is based on ResNet34 backbone and outputs confidence, angle, and distance. Out-of-distribution frames are rejected at ResNet Layer3.}
  \label{Overall Pipeline}
\end{figure}

The $\text{E}^2$VTS two-step text spotting system adopts Efficient and Accurate Scene Text Detector (EAST) as the text detector, and Convolutional Recurrent Neural Network (CRNN) as the text recognizer. The recognizer is connected with the detector via crop \& resize. A multi-stage image processor is proposed to further save the energy consumption. It has three stages, selecting the highest-quality frame in a sliding window, rejecting text-free images and cropping non-text regions, and rejecting out-of-distribution images. The pipeline is shown in Figure~\ref{Overall Pipeline},


\subsection{Revisiting RCNN: Crop \& Resize vs. Aligned RoI Pooling}

We compare two connecting mechanisms for the detector and the recognizer: crop+resize versus aligned RoI pooling. Examples of aligned RoI pooling include BezierAlign~\cite{Liu_2020_CVPR} for arbitrary-shaped text and RoIRotate~\cite{liu2018fots} for rotated text. 

Given the predicted bounding box from the detector, in crop+resize, the input to the recognizer is the cropped box area affinely transformed from the original image and resized to a fixed resolution. The detector and the recognizer are trained independently. In aligned RoI pooling, the input to the recognizer is the cropped box area affinely transformed from the feature map. The detector and the recognizer are trained jointly. Note that the text recognition loss uses the ground truth text regions instead of predicted text regions.

Unlike the benchmarks in image-based text spotting, real-world videos for text spotting are full of small size and poor quality text boxes. Consequently, crop+resize outperforms aligned RoI pooling for two reasons. First, aligned RoI pooling losses the discriminative details for small size text boxes due to the deep convolutions in the detector. In contrast, crop+resize enlarges the input resolution of small size text boxes and preserves their discriminative spatial details~\cite{wu2020context}. Second, text recognition (\ie, knowing what is the text) is intrinsically more difficult than text detection (\ie, knowing where is the text). Thus, feature sharing and joint training will lead to sub-optimal performance for both tasks~\cite{cheng2018revisiting,cheng2018decoupled}.

\subsection{Multi-Stage Image Processor}
Different from a single image, video frames are redundant and continuous in the temporal domain. Comparing one frame with its precedents and successors over certain metrics is a natural filtering process to select the most suitable frame for the later detection task. We also leverage sharp transitions of text regions to remove non-text background preliminarily to further boost efficiency. All these implementations are based on simple signal processing algorithms which are significantly faster than neural network models.


\subsubsection{Stage \rom{1}: Selecting the Highest-Quality Frame in a Sliding Window} 

\begin{figure*}[htb!]
  \centering
  \includegraphics[width=1.0\textwidth]{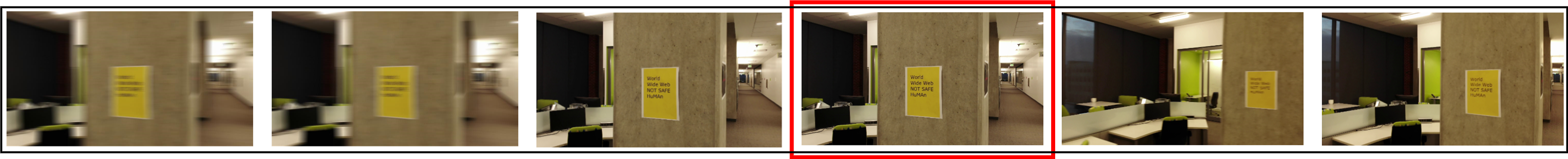}
  \caption{Sliding window for highest-quality frame selection. A window iterator is sliding over the temporally sub-sampled frames and quality scoring is conducted on the frames via the proposed measure. The highest ranked frame is selected.}
  \label{Stage I}
\end{figure*}

\paragraph{Problem Definition:} Blur is the major artifact in UAV captured videos due to camera shake, depth variation, object motion or a combination of them~\cite{wu2020david,kupyn2019deblurgan}. Among all the frames describing the same visual scene, the clearest image gives the least amount of detector or recognition error. Since blurred frames contain less energy in the high frequency components, in their associated power spectrum~\cite{yi2016lbp}, the power tends to fall much faster with increasing frequency, compared with clear frames. Therefore, the average of the power spectra of clear frames is higher than these degraded ones, as degraded ones have a steeper slope on their power spectrum. 

\paragraph{Implementation:} In Fig.~\ref{Stage I}, we propose a sliding window mechanism and select the highest-quality frame in each window. Given a video containing $L$ frames, the $i$-th window $\mathcal{W}_i$ is obtained via:
\begin{equation}
    \mathcal{W}_i = S_{(i, N)}({I_1,...,I_L}),
\end{equation}
where $S$ represents the sliding rule and $N$ is the window size. The selected highest-quality frame in $\mathcal{W}_i$ is:
\begin{equation}
    I_{HQ} = \argmax_{I\in W_i} \mathcal{G}(I),
\end{equation}
where $\mathcal{G}$ is the quality measure. We propose two measures in this work: variance of Laplacian\cite{pech2000diatom} and average fast Fourier transform (FFT) magnitude defined as:
\begin{equation}
    \begin{split}
        \mathcal{G}_{FFT} &=\frac{1}{hw} \lVert \text{FFT}(I_0) \rVert, \\
        \mathcal{G}_{LV} &= \text{Var}(k_L * I_0),
    \end{split}
\end{equation}
where $I_0$ is a given frame with height $h$ and width $w$.
FFT magnitude measure is an approximation of power spectrum density in frequency domain. Variance of Laplacian stresses spatial information by counting sharp transitions in the frame. These two measure works in a complementary way. Therefore, we integrate the two methods by taking a weighted average of two measures' scores ranking over a certain window. Let $\text{rank}(I,W,\mathcal{G})$ denotes a function that returns the rank of frame $I$ among all the frames in the window $W$ scored by the quality measure $\mathcal{G}$ in ascending order. The selected highest-quality frame is:  
\begin{align}
    \begin{split}
        I_{HQ} = \argmax_{I\in W_i} [&\lambda \cdot \text{rank}(I, W_i, \mathcal{G}_{FFT}) \\
            + &(1-\lambda) \cdot \text{rank}(I, W_i, \mathcal{G}_{LV})],
    \end{split}
\end{align}
where $\lambda$ is the relative weight parameter.

In practice, the video sequence is sub-sampled at rate $r$ to further boost efficiency before applying the sliding window filter. As a hyper-parameter, the sub-sample rate $r$ has a great impact on the tradeoff between energy-efficiency and performance. Although setting higher sub-sample rate could save more energy, it has higher chance to miss scenes for text spotting.






\subsubsection{Stage \rom{2}: Rejecting Text-free Images and Cropping Non-Text Regions} \label{reject non-text image}

\begin{figure*}[t]
  \centering
  \includegraphics[width=1.0\linewidth]{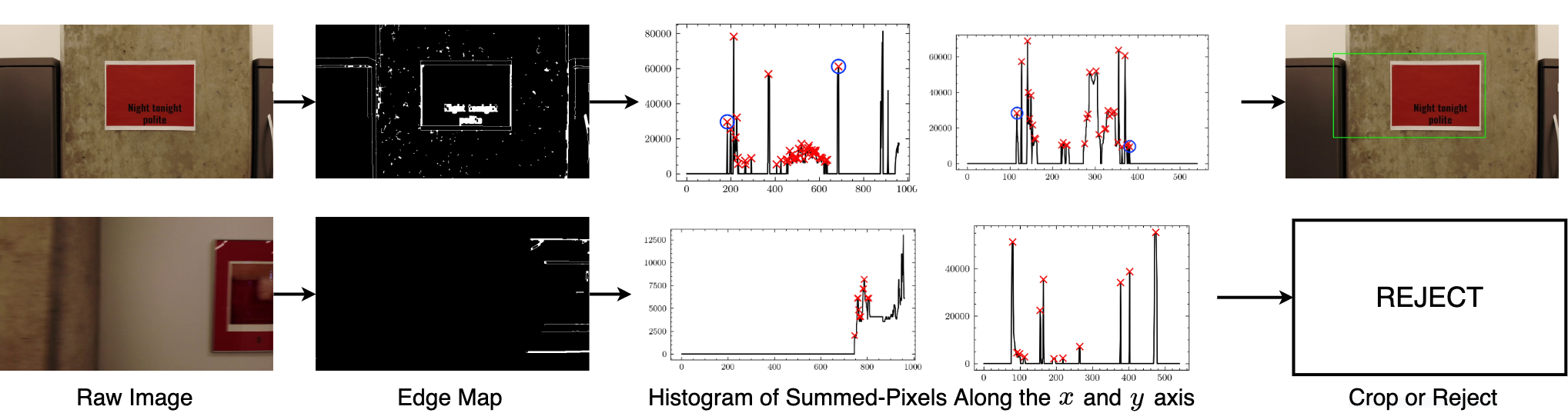}
  \caption{Cropping text foreground: we use the histogram to analyze the edge information of the selected frame. If the number of peaks and the mean of intensity satisfy predefined thresholds, text bounding coordinates will be selected from peaks info. Otherwise, the frame will be discarded.}
  
  \label{Stage_II}
\end{figure*}
\paragraph{Problem Definition:}
Known for high time complexity and energy consumption, connected component-based text detection depends on maximally stable extremal region (MSER) as character candidates, and stroke width transform (SWT) for filtering and pairing of connected components. Given the observation that cohesive characters compose a word or sentence sharing similar properties such as spatial location, size, and stroke width, we turn to Canny edge detector~\cite{canny1986computational} to locate the edge pixels that build the text's structure (a.k.a. contour). 
\paragraph{Implementation:}
In Fig.~\ref{Stage_II}, we further reject text-free images and crop non-text regions. First, Canny edge detector is applied on the three channels of the input image $I_{yuv}$ represented in YUV color space, and the three channels $(Y_c, U_c, V_c)$ are merged by bitwise OR (denoted as ``$\BitOr$'') operation to obtain the edge map $I_e$. 
Then, morphological closing is applied on the $I_e$ to remove small holes and merge connected components. If any text region is present in the image, a binary image with continued text characters will be returned. 
Next, the histogram map is obtained by summing up pixels along the $x$ and $y$ axis~\footnote{With the origin in lower left corner, the $x$-axis is running from left to right and the $y$-axis is running from bottom to up.}:
\begin{align} \label{Histogram}
    H_{x}[i] = \displaystyle\sum_{k=1}^h I_c[i,k], 
    H_{y}[j] = \displaystyle\sum_{k=1}^w I_c[k,j],
\end{align}
where $w$ and $h$ are the width and height of the image.
After that, all the peaks~\footnote{The peaks are all local maxima by comparing neighboring values in the histogram.} for these two histogram maps, \ie, $P_x$ and $P_y$, are found. Text regions are assumed to fall within the peaks. 
Finally, Text-free images are rejected based on two preset thresholds ($\theta, \alpha$) on the peak intensities and numbers, respectively. Note that the second-stage selector cannot deal with images with complicated backgrounds, since the peaks value varies along both axes without any identifiable pattern. Therefore, images with complicated background whose peaks are consistently high along the entire $x$ and $y$ axis are accepted.

Cropping text regions improves the SNR~\footnote{We treat text related pixels as signal and all other pixels as noise.} in the image. On images with simple background, the text regions are assumed to lie between ($x_l$, $y_b$) and ($x_r$, $y_t$). The coordinates of the text region are obtained from the peaks via
\begin{align}
    x_l, x_r, y_b, y_t = P_x[1], P_x[-1], P_y[1], P_y[-1].
\end{align}
The details of the second-stage selector is shown in Algorithm~\ref{2nd_stage_selector}. 
\begin{algorithm}
\caption{Rejecting Text-free Images or Cropping Text Regions}\label{2nd_stage_selector}
Initialization: $\theta$, $\alpha$: predefined thresholds\\
\SetAlgoLined
    $I_{yuv} \gets \text{RGB2YUV}(I)$ \\
    $Y_c, U_c, V_c \gets \text{CannyEdge}(I_{yuv})$ \\
    $I_e \gets Y_c \BitOr U_c \BitOr V_c$ \\
    $I_c \gets \text{MorphClose} (I_e)$ \\
    $H_x, H_y \gets \text{Histogram}(I_c)$ \tcp{sum up pixels among axis}
    $P_x, P_y \gets \text{FindPeaks}(H_x, H_y)$ \\
    $\mu_x, \mu_y \gets \text{Mean}(P_x), \text{Mean}(P_y)$ \\
    \tcp{Whether the number of peaks or the mean of intensity is less than preset thresholds}
    \If{$\text{Count}(P_x) \leq \theta$ \text{or} $\text{Count}(P_y) \leq \theta $ \text{or} $\mu_x \leq \alpha$ \text{or} $\mu_y \leq \alpha$} 
    {
    REJECT
    }
    \Else{
    ACCEPT \\
    $x_l,x_r,y_b,y_t \gets P_x[1],P_x[-1],P_y[1],P_y[-1]$ \\
    \textbf{return} $I[x_l:x_r, y_b:y_t]$ \\
    }
\end{algorithm}



\subsubsection{Stage \rom{3}: Rejecting Out-of-Distribution Images}
\begin{figure*}[htb!]
  \centering
  \includegraphics[width=1.0\textwidth]{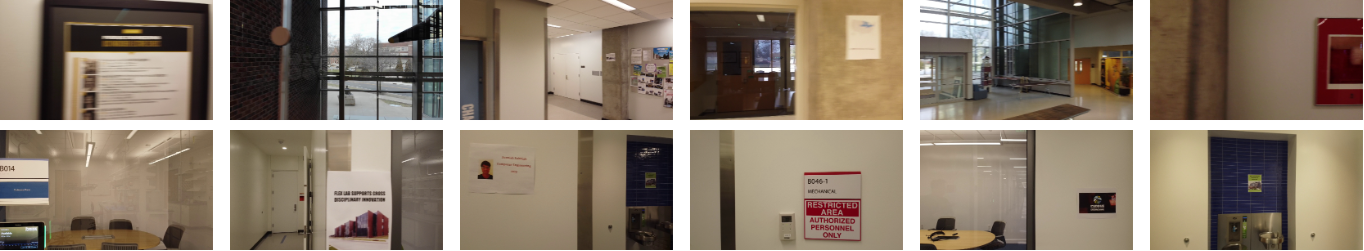}
  \caption{The negative samples in the first row and positive samples in the second row are used to train the out-of-distribution rejector. Heavily-blurred, text-free, are truncated-text are all considered as negative cases to be rejected.}
  \label{fig:ODR data}
\end{figure*}

\paragraph{Problem Definition:} A ``trained'' $\text{E}^2$VTS model $f$ with fixed parameters is able to fit a distribution $\mathcal{X}_f$ defined on the image space. During inference, rejecting the out-of-distribution images in an early-exit way could greatly reduce energy consumption.
The out-of-distribution rejection problem~\cite{liang2018enhancing} can be formulated as a binary classification. Examples of the positive cases and negative cases used to train the rejector are shown in Fig.~\ref{fig:ODR data}. 

\paragraph{Implementation:}
 Grad-CAM~\cite{grad-cam}, a visual explanations technique via gradient-based localization, is deployed to locate the first text semantic-aware layer $l$ for our model. The outputs of the text semantic-aware layer $H_l$ serve as the high-level features to distinguish the out-of-distribution images from the in-distribution ones. Support Vector Machines (SVM) is used for binary classification on $H_l$. SVM is preferred over the deep model due to its small size in the number of parameters and low latency. 

\section{Experiments}

\subsection{Experiment Settings}
\begin{figure}[htb!]
\setlength{\tabcolsep}{0em}
\begin{tabular}{llll}
	\includegraphics[width=0.115\textwidth]{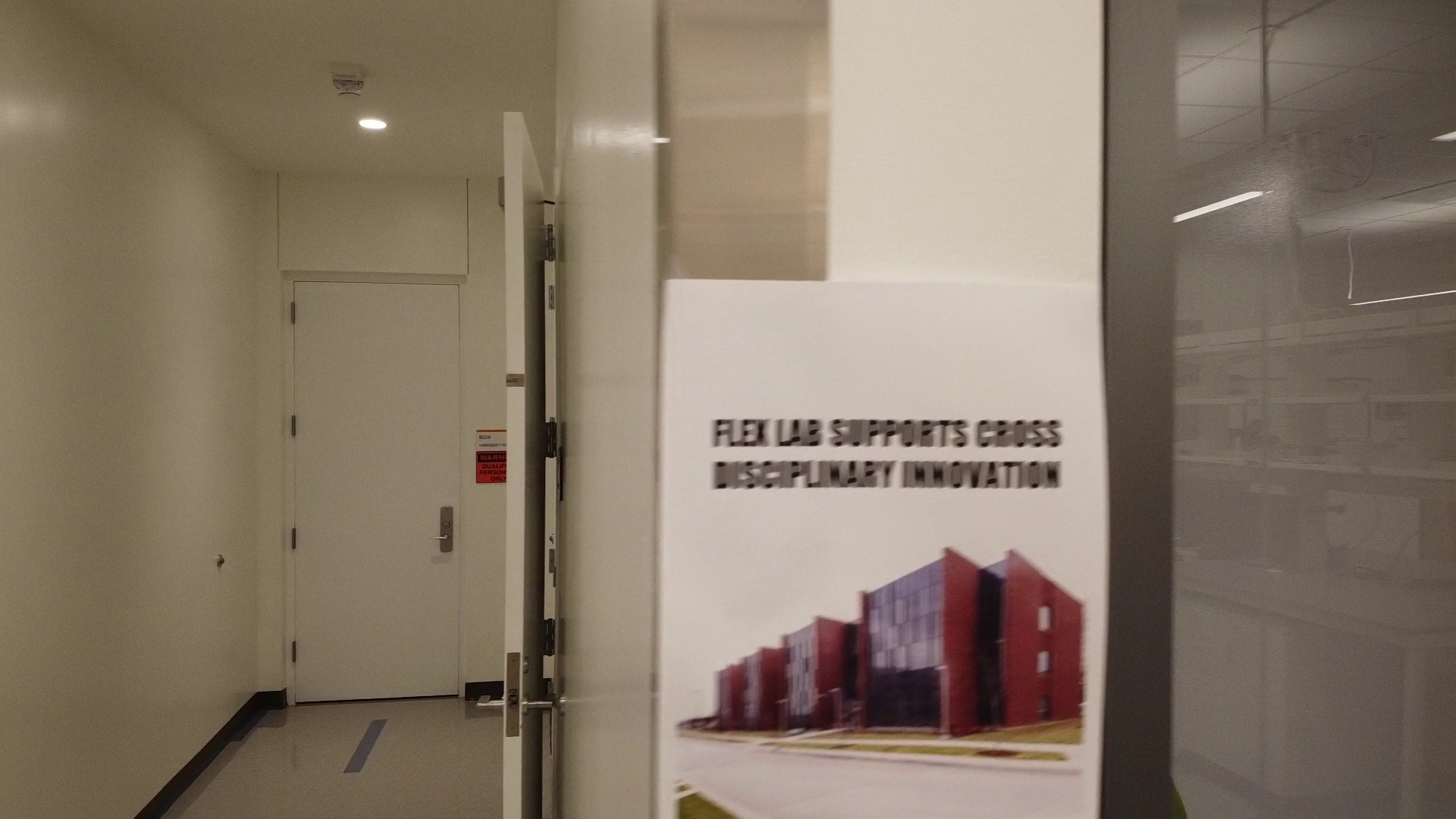}
	\includegraphics[width=0.115\textwidth]{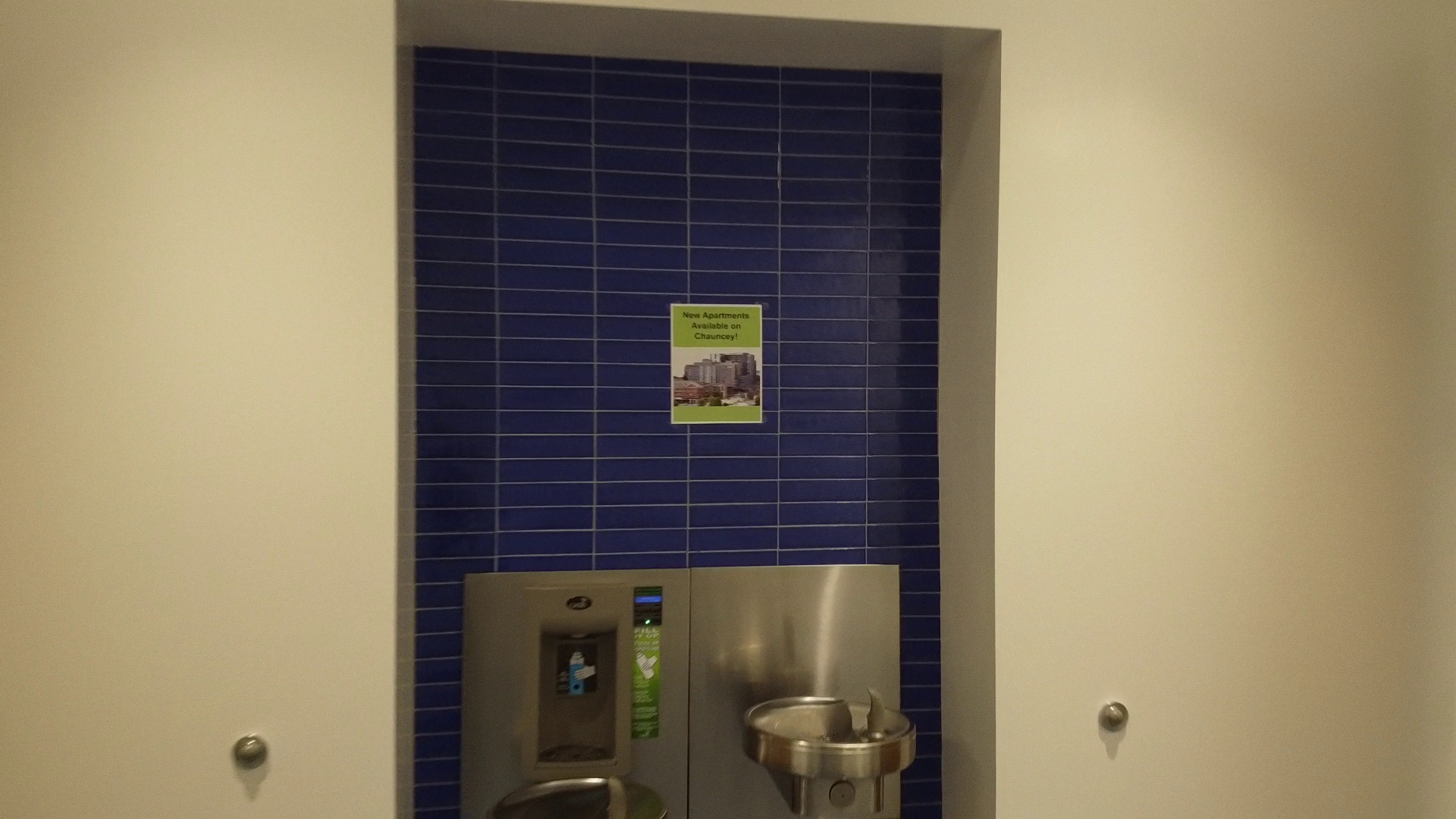}
	\includegraphics[width=0.115\textwidth]{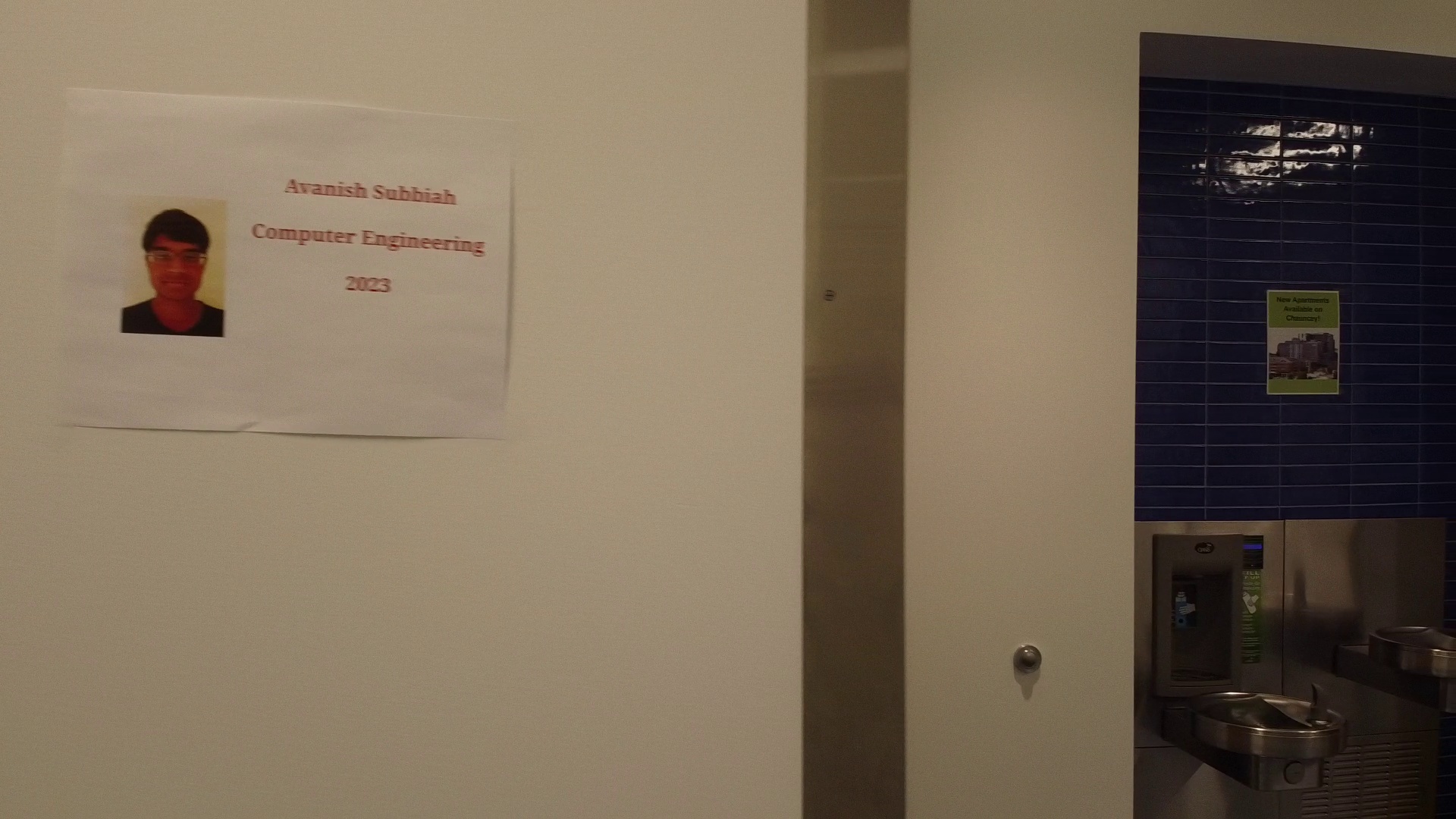}
	\includegraphics[width=0.115\textwidth]{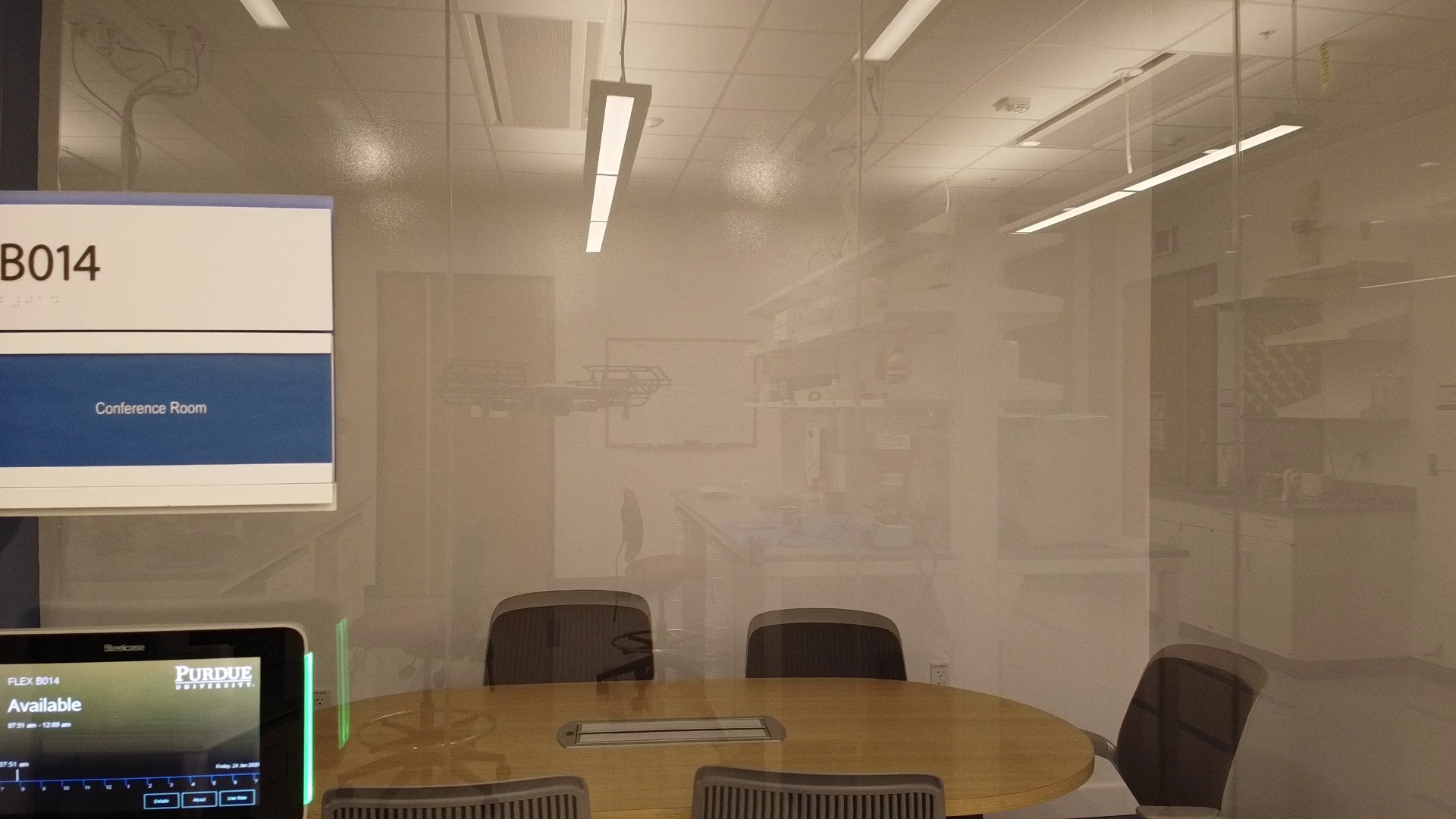}\\
\end{tabular}

\begin{tabular}{llll}
	\includegraphics[width=0.115\textwidth]{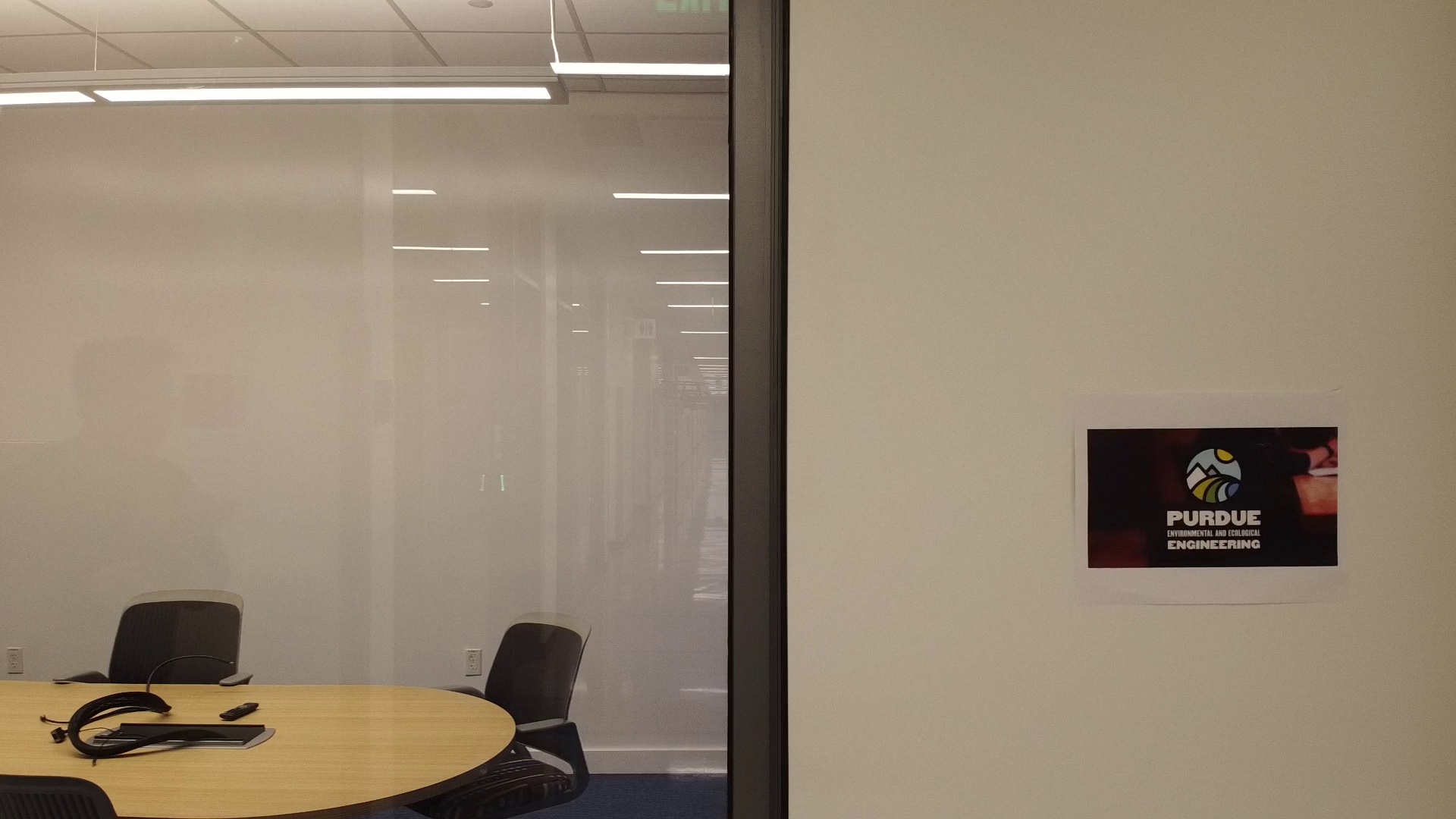}
	\includegraphics[width=0.115\textwidth]{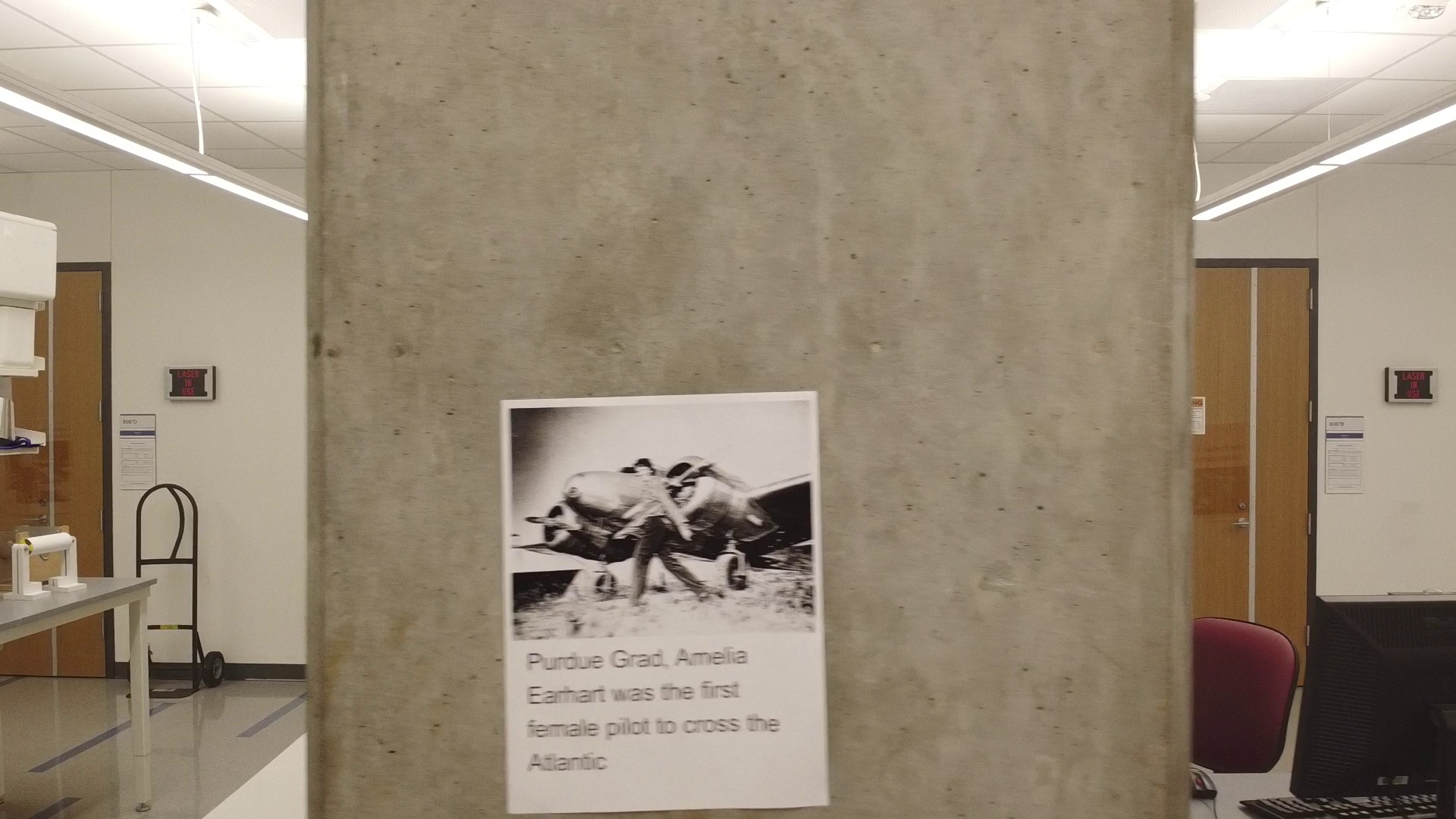}
	\includegraphics[width=0.115\textwidth]{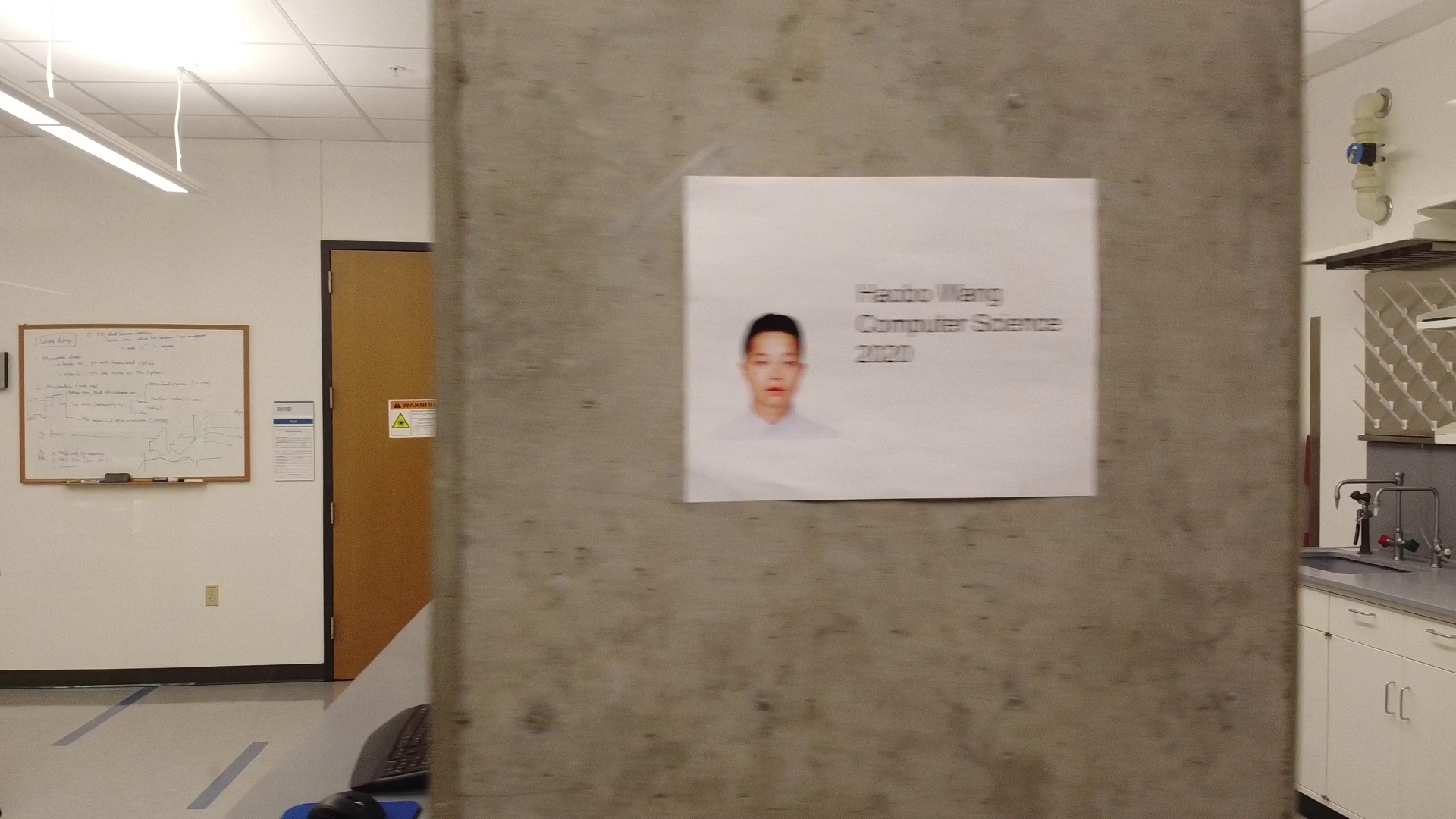}
	\includegraphics[width=0.115\textwidth]{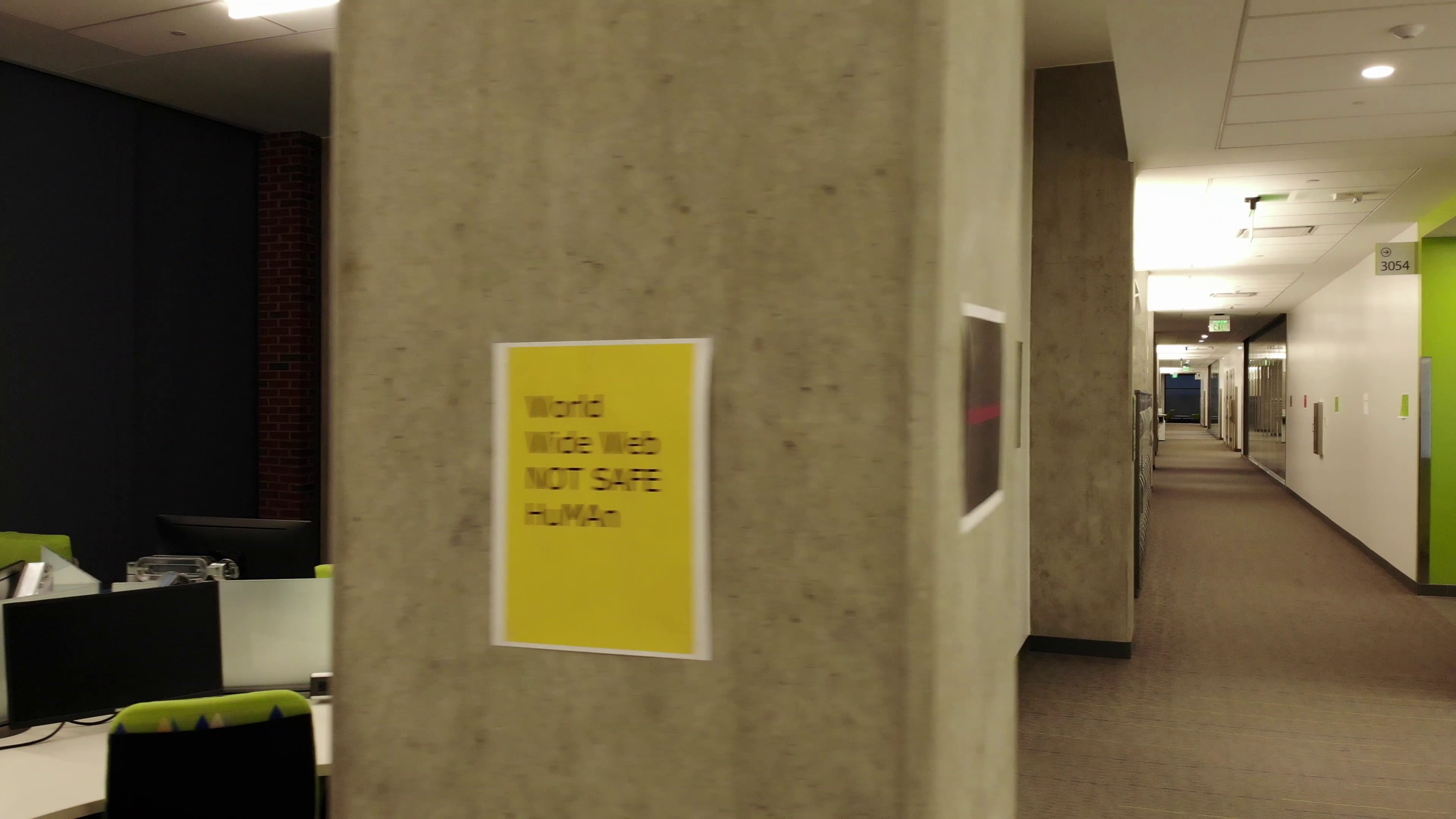}
	\label{fig:stb}
\end{tabular}
\caption{Sample Images from the LPCVC-20 Video Text Spotting Dataset.}
\label{fig:lpcv_sample}
\end{figure}

\subsubsection{Datasets and Evaluation Protocols} %
We evaluate the proposed $\text{E}^2$VTS approach on the LPCVC-$20$ video text spotting dataset, abbreviated as \textbf{LPCVC-20}. The videos are captured by UAVs flying indoors on the corridors, where tons of posters and board signs with rotated text are presented. Five videos were used for training and one video was reserved for testing. After decomposing videos into frames, we handpick text-related images as our system experiment datasets which include $7,886 $ for training and $2,033$ for texting. Furthermore, the text was annotated using the Auto Labeling algorithm described in the coming subsection. LPCVC-20 consists of images of resolution $3840 \times 2160$, $1920 \times 1080$, and $1280 \times 720$.

IoU, IoP, IoG~\footnote{Given a ground truth bounding box area G and predicted bounding box area P the IoU is $(P \cap G) / (P \cup G)$, IoP (a.k.a precision) is $(P \cap G) / P$, and IoG (a.k.a recall) is $(P \cap G) / G$.} are used for detection and edit-distance is used for recognition. For each ground truth, the predicted bounding box with the maximum IoU is selected and the edit distance is calculated between the ground truth text label and the predicted text.

\paragraph{Auto Labeling: Image Registration-Aided Annotation for Video Text Spotting}
\begin{figure*}[htb!]
  \centering
  \includegraphics[width=1.0\textwidth]{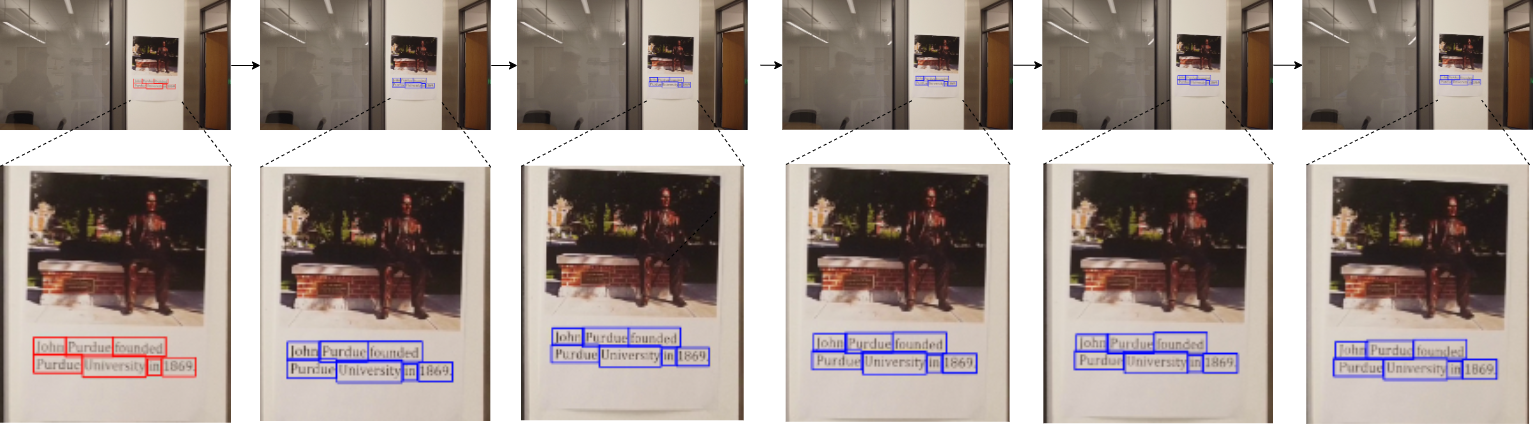}
  \caption{Qualitative Results of Auto Labeling: extract features from the source frame and the target frame. Conduct feature matching and perspective transform to update bounding boxes annotation. Repeat the process until the end of the scene.}
  \label{fig:auto_label}
\end{figure*}

Given the observation that temporally consecutive frames are describing the same scene, we propose Auto-Labelling in Algorithm~\ref{algo:auto_labeling} to aid the video annotation, which utilizes the videos' temporal redundancy and continuity. It takes advantage of feature matching and perspective transformation to transfer the annotated bounding box from the source frame to the target frame. Figure~\ref{fig:auto_label} shows the annotation results produced by Algorithm~\ref{algo:auto_labeling}.

\begin{algorithm}
\caption{Auto-Labeling}\label{algo:auto_labeling}
\SetAlgoLined
\tcp{Annotate the 1st frame} 
    $b_s \gets \text{Annotate}(V[1])$ 
    \tcp{ Number of frames describing the same scene}
    $N \gets \text{Size}(V)$ \\
    \For {$i \gets 2$ \KwTo $N$} {
    \tcp{Next Adjacent frame}
	    $I_t \gets V[i]$ 
	    \tcp{Feature Matching}
	    $k_1, d_1 \gets {\small \text{SIFT}(I_s); k_2, d_2 \gets \text{SIFT}(I_t)}$ \\
	    $m \gets {\small \text{LoweRatioTest}(\text{BFMatcher}(d_1, d_2))}$ \\
	    $p_s, p_t \gets {\small \text{FilterKeyPts}(m, k_1), \text{FilterKeyPts}(m, k_2)}$ \\
	    \tcp{Perspective Transformation}
	    $b_t \gets {\small \text{Perspective}(b_s, \text{HomographyMatrix}(p_s,p_t))}$ \\
		$I_s \gets I_t; b_s \gets b_t$
	}
\end{algorithm}

\paragraph{Energy Consumption Measurement}
A USB power meter~\footnote{MakerHawk UM34C USB 3.0 Multimeter Bluetooth USB Voltmeter Ammeter} is used to measure energy consumption. We connect the USB power meter in series to the power supply of the Raspberry Pi. With this setup, the power meter can real-time measure the current through the Raspberry Pi. Since the voltage for Raspberry Pi is constantly 5V, we can calculate the energy consumption by recording the current values. The power meter is connected with a computer through Bluetooth and the energy measurements of the Raspberry Pi are recorded using \cite{rdusbgithub}. The timestamps for model inference are written down to measure the latency for the model.

\subsubsection{Model Compression}
\paragraph{Pruning}
Pruning algorithm compresses neural network by removing redundant weights or channels of layers. For a Raspberry Pi, structured pruning is preferred over unstructured pruning, since structured pruning does not require specific hardware support for deployment. For our experiment, we applied $\ell_1$ filter Pruner with a one-shot pruning strategy and a sparsity rate of $0.7$, which allowed our model to achieve the best trade-off between accuracy and energy efficiency.

\paragraph{Quantization}
Quantization refers to techniques for using a reduced precision integer representation for weights and activations. For Raspberry Pi, Pytorch provides QNNPACK backends which supports running quantized operators efficiently on ARMS CPU. For our experiment, we applied static post quantization on all convolutional and fully-connected layers; and applied dynamic post quantization on the LSTM modules in the CRNN model. 






\subsection{Ablation Studies}
\subsubsection{Crop \& Resize vs. Aligned RoI Pooling}
In this section, we conduct ablation studies for the two-step Crop and Resize $\text{E}^2$VTS text spotting model and the two-stage Aligned RoIPool text spotting model. From Tables ~\ref{tab:east_crnn_dir} and ~\ref{tab:east_crnn_dfr} it can be seen that the $\text{E}^2$VTS model performs better than the Aligned RoIPool model at all resolutions. The different factors that influence the performance of the model are also measured. From Tables ~\ref{tab:east_crnn_dir} and ~\ref{tab:east_crnn_dfr} it can be concluded that the greater bounding box to character count ratio and the lesser character count improves the recognition performance. Table ~\ref{tab:model_deployment_results} shows the deployment results of  $\text{E}^2$VTS on Raspberry Pi.

\begin{table}
\centering
\caption{$\text{E}^2$VTS results on LPCVC-20}
\begin{tabular}{c|c|c|c|c}
\arrayrulecolor{black}
\toprule
\multicolumn{2}{c|}{EditDistance \textbackslash{} Resolution} & 1200 & 600 & 300 \\ 
\hline
\multirow{3}{*}{BBox Area/Char Count} & \textless{}=20 & N/A & 6.86 & 3.69 \\ 
\cline{2-5} & \textless{}=60 & 5.25 & 2.12 & 3.60 \\ 
\cline{2-5} & \textgreater{}60 & 1.93 & 2.00 & 2.92 \\ 
\hline
\multirow{3}{*}{Char Count} & \textless{}=4 & 1.08 & 1.31 & 2.21 \\ 
\cline{2-5} & \textless{}=8 & 1.86 & 2.08 & 3.56 \\ 
\cline{2-5} & \textgreater{}8 & 4.20 & 3.79 & 5.32 \\ 
\hline
\multicolumn{2}{c|}{Total} & 1.93 & 2.04 & 3.26 \\
\hline
\end{tabular}
\label{tab:east_crnn_dir}
\end{table}

\begin{table}
\centering
\caption{FOTS results on LPCVC-20}
\begin{tabular}{c|c|c|c|c} 
\hline
\multicolumn{2}{c|}{EditDistance \textbackslash{} Resolution} & 1200 & 600 & 300 \\ 
\hline
\multirow{3}{*}{BBox Area/Char Count} & \textless{}=20 & N/A & 6.84 & 4.19 \\ 
\cline{2-5} & \textless{}=60 & 4.56 & 2.84 & 5.19 \\ 
\cline{2-5} & \textgreater{}60 & 2.62 & 3.48 & 5.97 \\ 
\hline
\multirow{3}{*}{Char Count} & \textless{}=4 & 1.83 & 2.38 & 3.28 \\ 
\cline{2-5} & \textless{}=8 & 2.98 & 4.21 & 6.12 \\ 
\cline{2-5} & \textgreater{}8  & 3.95 & 4.92 & 8.61 \\ 
\hline
\multicolumn{2}{c|}{Total} & 2.65 & 3.48 & 5.25 \\
\arrayrulecolor{black}
\bottomrule
\end{tabular}
\label{tab:east_crnn_dfr}
\end{table}


\begin{table}

\centering
\caption{Performance, Latency, and Energy Measurement of  $\text{E}^2$VTS on Raspberry Pi}
\adjustbox{max width=\columnwidth}{%
\begin{tabular}{*{7}{c}} 
\arrayrulecolor{black}
\toprule
Model           & IoU & IoP &IoG& EditDistance & Latency & Avg Energy \\
\midrule
$\text{E}^2$VTS & 72.21 & 76.24 & 93.94 & 1.39 & 12.90 & 31.77 \\
\bottomrule
\end{tabular}}
\label{tab:model_deployment_results}
\
\end{table}

\subsubsection{Multi-Stage Image Processor}
We compare the overall performance for our method after incorporating different data level efficiency techniques. As shown in Table. \ref{tab:eval_data_efficiency}, incorporating data level efficiency results in a better performance in both accuracy and efficiency. 

\begin{table}[htb!]
\setlength{\tabcolsep}{3pt}
\caption{Ablation studies on the multi-stage image processor. Performance, latency, and energy consumption are evaluated.}
\centering
\adjustbox{max width=\columnwidth}{%
\begin{tabular}{cccccc}
\toprule
Stage \rom{1} & Stage \rom{2} & Stage \rom{3} & Latency & Energy & EditDistance \\
\midrule
$\checkmark$ & {} & {} & {627.43} & {1841.49} & {\textbf{0.78}} \\
{} & $\checkmark$ & {} & {545.20} & {1349.23} & {1.14} \\
{} & {} & $\checkmark$ & {571.48} & {1482.31} & {1.05} \\
$\checkmark$ & $\checkmark$ & $\checkmark$ & {\textbf{528.12}} & {\textbf{1267.2}} & {0.96} \\
\arrayrulecolor{black}
\bottomrule
\end{tabular}}
\label{tab:eval_data_efficiency}
\end{table}

Based on the results in Table~\ref{tab:eval_data_efficiency}, Stage I data pre-processing greatly improves the accuracy of the model by selecting the best quality frame within a window size as the model's input. The latency and energy decrease slightly due to the extra cost introduced by quality scoring and the decrease of the video's sub-sample rate. Stage II data pre-processing mainly decreases the latency and average energy consumption of the model by improving the SNR in the image and rejecting low-quality and non-text frames. Stage III data pre-processing also decreases latency and energy consumption by rejecting out-of-distribution frames at an early stage of the detection model. The integration of Stage I, II, and III data pre-processing benefits the model from the perspective of speed and energy consumption. 

\subsubsection{Deployment on Raspberry Pi}
In this section, we evaluate the overall performance of our method after incorporating different model level efficiency techniques, which include pruning and quantization.

\begin{table}[htb!]
\setlength{\tabcolsep}{3pt}
\caption{Ablation studies on pruning ($\mathbf{P}$) and quantization ($\mathbf{Q}$).}
\label{table_model_level_eff}
\centering
\scalebox{1}{
\begin{tabular}{ccccc}
\arrayrulecolor{black}
\toprule
$\mathbf{P}$ & $\mathbf{Q}$ & Latency & Energy & EditDistance \\
\midrule
{} & \checkmark & {76.48} & {195.25} & {\textbf{1.09}} \\
$\checkmark$ & {} & {56.67} & {164.73} & {1.12} \\
$\checkmark$ & $\checkmark$ & {\textbf{12.90}} & {\textbf{39.23}} & {1.14} \\
\arrayrulecolor{black}
\bottomrule
\end{tabular}}
\label{tab:eval_model_efficiency}
\end{table}

Based on the results in Table~\ref{tab:eval_model_efficiency}, model pruning and quantization significantly decrease latency and average energy consumption respectively. Although implementing model compression results in a sightly drop in accuracy, the tradeoff between energy efficiency and accuracy shows that incorporating model level efficiency notably boosts overall performance.




\section{Conclusion} 
In this paper, we proposed an energy-efficient video text spotting solution, dubbed as $\text{E}^2$VTS, for Unmanned Aerial Vehicles. $\text{E}^2$VTS is an energy-efficiency driven model without compromising text spotting performance. The proposed system not only utilizes data level efficiency enhancement techniques but also makes use of model level efficiency boosting methods such as pruning and quantization. Specifically, a sliding window is used to select scene-wise highest quality frame; a Canny edge based algorithm is proposed to reject text-free images and non-text frames; a dynamic routing mechanism emphasizes the in-distribution inputs. Far from the application on UAV devices, our video text spotting system is competent for any energy-constrained scenario. 
\section{Acknowledgement}
We would like to express our sincere gratitude to Ytech Seattle AI Lab, FeDA Lab, Kwai Inc., for their generous financial and technical supports during our participation in the LPCVC 2020 UAV Video Track.

\end{document}